\documentclass[letterpaper, 10 pt, conference]{ieeeconf}
\IEEEoverridecommandlockouts
\usepackage[utf8]{inputenc}
\usepackage[T1]{fontenc}

\usepackage{hyperref}
\usepackage{amsmath}
\usepackage{amssymb}
\usepackage{multirow}
\usepackage{booktabs}
\usepackage{graphicx}
\usepackage{caption}
\usepackage{subcaption}
\usepackage{algorithm}
\usepackage{algorithmic}
\usepackage{colortbl}
\usepackage{xcolor}
\usepackage{float}
\let\labelindent\relax
\usepackage{enumitem}

\newcommand{\agentslabench}{\textsc{AgentSLABench}}
\newcommand{\easr}{\textsc{EASR}}

\title{\LARGE \bf AgentSLABench: Evaluating and Benchmarking Agentic Systems Under Resource Constraints}

\author{ \parbox{3 in}{ \centering Meher Bhaskar Madiraju
        {\tt\small meherbhaskar.madiraju@gatech.edu}}
        \hspace*{ 0.5 in}
        \parbox{3 in}{\centering Meher Sai Preetam Madiraju
        {\tt\small mehersaipreetam@gatech.edu}}
}

\begin{document}

\maketitle
\thispagestyle{empty}
\pagestyle{empty}

\begin{abstract}
We present \agentslabench{}, a resource-aware evaluation framework for autonomous AI agents that measures \textbf{correctness alongside latency, cost, compute, memory, and network usage} under declared resource budgets. Unlike standard benchmarks that report only accuracy, \agentslabench{} produces a \textbf{multi-dimensional profile} per agent per task --- the same way systems profilers (\texttt{perf}, \texttt{pprof}, \texttt{cProfile}) measure resource consumption of code, but extended with task correctness as a first-class dimension. \agentslabench{} provides 16 task environments across 6 categories (5 core: multi-hop QA, retail substitution, code generation, web shopping, travel planning; 11 extended) with isolated Docker containers, declared CPU/memory/time/network budgets, sealed test sets with SHA256 hashes, and a standardized profiling protocol. We profile 5 general-purpose baseline agents (ReAct, PlanAndSolve, Reflexion, CoT, Random) plus 4 task-specialized agents, finding that specialized agents achieve 100\% success on 3/5 core tasks (fact\_qa, web\_shopping, travel\_planning) and 66.7--83.3\% on retail and code\_gen, while \textbf{general baselines fail entirely on 4/5 domain tasks}. Crucially, we report the \textbf{Efficiency-Adjusted Success Rate (\easr{})} --- success weighted by resource consumption relative to declared budgets --- revealing that high accuracy at unbounded cost is not production-viable. We release the full infrastructure, sealed test sets, and profiling results to enable reproducible, resource-aware agent evaluation.
\end{abstract}


\section{Introduction}
\label{sec:intro}

Autonomous agents powered by large language models (LLMs) are transitioning rapidly from research prototypes to production systems --- deployed in customer support, software engineering, financial analysis, and enterprise automation. Yet the evaluation methodologies remain stuck in a pre-production mindset: \textbf{``Did it succeed?''} dominates leaderboards and papers, while \textbf{``Can it run in production?''} is systematically ignored.

An agent achieving 90\% task accuracy but costing \$50/query and taking 5 minutes is useless for real-time retail substitution where decisions must complete in $<200$ms at $<\$0.01$. Similarly, an agent that solves coding tasks but requires 32GB GPU memory and 10 API calls per step cannot be deployed in cost-sensitive environments. Standard benchmarks (WebShop~\cite{yao2022webshop}, ALFWorld~\cite{shridhar2020alfworld}, SWE-bench~\cite{jimenez2023swebench}, GAIA~\cite{mialon2023gaia}, AgentBench~\cite{liu2023agentbench}, ToolBench~\cite{xu2023toolbench}) measure task accuracy in unconstrained environments with effectively unlimited compute, memory, and API budgets. Systems profilers (\texttt{perf}, \texttt{pprof}, \texttt{cProfile}, \texttt{nsys}, VTune) measure CPU, memory, cache, and GPU utilization of \textit{code functions} but have no notion of task correctness. \textbf{No existing framework measures both task success and resource consumption under production-realistic constraints.}

This gap is not merely academic. Production deployment of LLM agents faces hard constraints that current evaluations ignore:
\begin{itemize}
    \item \textbf{Latency SLAs:} Often $<500$ms for user-facing, $<2$s for backend workflows
    \item \textbf{Cost budgets:} Cents per query, not dollars (e.g., \$0.01--\$0.10 per resolution)
    \item \textbf{Memory limits:} Container quotas (typically 2--8GB for CPU workloads)
    \item \textbf{Rate limits:} External API throttling (e.g., 60 req/min on search APIs)
    \item \textbf{Safety/compliance:} PII leakage, unauthorized tool calls, budget overruns
\end{itemize}
An evaluation that ignores these dimensions produces ``leaderboard champions'' that fail in deployment. Industry surveys confirm this: teams report that agents passing accuracy benchmarks routinely violate latency and cost SLAs in staging, requiring costly re-engineering~\cite{langsmith2024production}.

Consider a concrete example: a retail substitution agent deployed at a major e-commerce platform must recommend alternative products when items go out of stock. The production SLA requires P99 latency $<200$ms, cost $<\$0.005$ per request, and memory $<512$MB. A ReAct-based agent using GPT-4 achieves 78\% substitution accuracy in unconstrained evaluation but averages 3.2s latency, \$0.12 cost, and 1.2GB memory --- violating all three SLAs. A specialized RetailAgent using a smaller model with cached product embeddings achieves 83\% accuracy at 142ms, \$0.002, and 180MB --- meeting all constraints. Current benchmarks would report only the 78\% vs 83\% accuracy, missing the critical deployment reality.

We introduce \textbf{agent profiling}: extending systems profiling to autonomous agents with \textit{correctness as a measured dimension alongside latency, cost, compute, memory, and network}. The output of \agentslabench{} is not a scalar ``score'' but a \textbf{profile} --- a structured record of (success, latency, cost, peak memory, CPU time, network calls, safety violations) per episode, under declared resource budgets. This mirrors how systems engineers profile code: they do not ask ``is this function correct?'' but rather ``what is its performance profile?'' --- then optimize within constraints.

\begin{figure*}[t]
\centering
\begin{minipage}{0.95\linewidth}
\centering
\includegraphics[width=\linewidth]{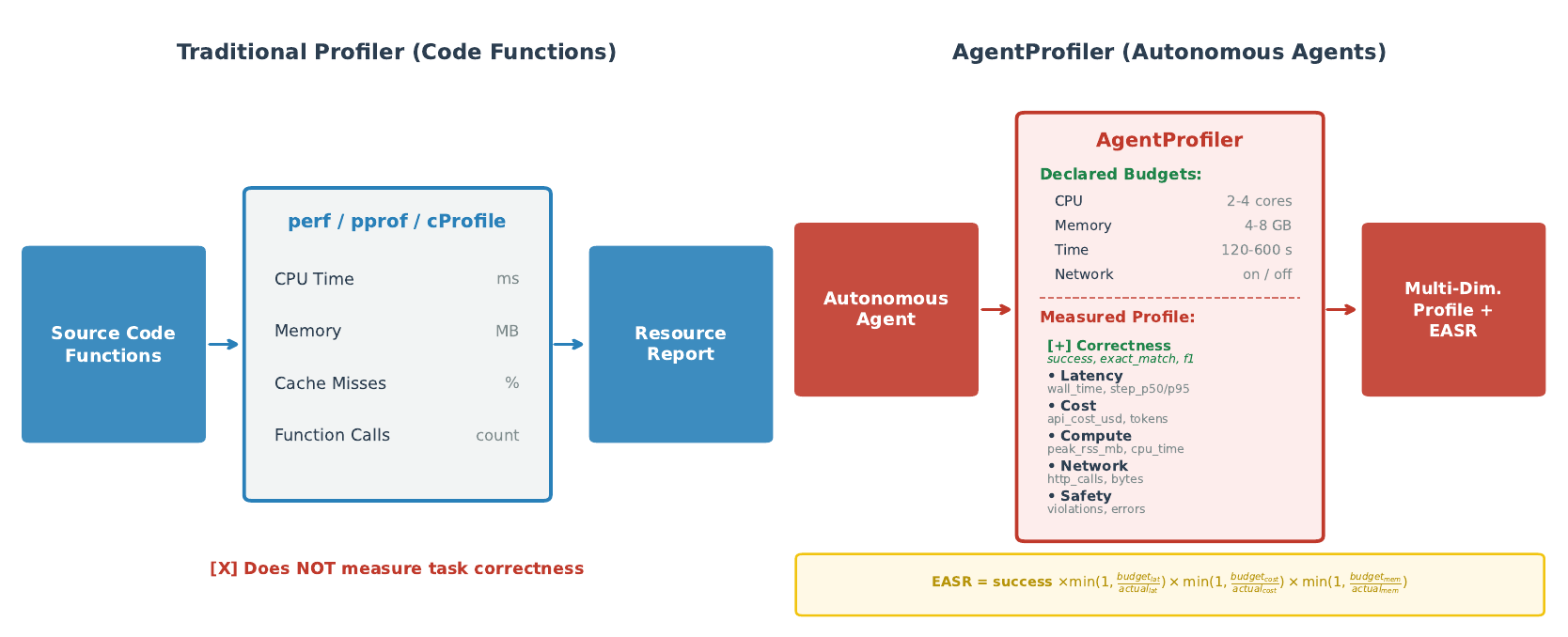}
\caption{\textbf{Standard profiler vs. AgentSLABench.} (Left) Traditional profilers measure resource usage of code functions without task correctness. (Right) \agentslabench{} extends this to autonomous agents, adding task correctness as a first-class dimension alongside latency, cost, compute, memory, and network.}
\label{fig:profile_concept}
\end{minipage}
\end{figure*}

\textbf{Contributions:}
\begin{enumerate}
    \item \textbf{Resource budgets as first-class specifications:} Each task declares CPU, memory, wall-time, and network budgets --- agents are evaluated within these envelopes via Docker-enforced limits.
    \item \textbf{Efficiency-Adjusted Success Rate (\easr{}):} A metric rewarding success \textit{within} declared budgets (Eq.~1), distinguishing ``success at any cost'' from ``production-viable success.''
    \item \textbf{Profile output format:} Per-episode JSONL with all six dimensions (correctness, latency, cost, compute, network, safety) enabling Pareto analysis and multi-objective optimization.
    \item \textbf{Reproducible infrastructure:} 16 Docker-isolated tasks, sealed test sets with SHA256 commitments, multi-seed profiling (3+ seeds), versioned container images.
    \item \textbf{Empirical finding:} Specialized agents achieve 100\% success on 3/5 core tasks and 66.7--83.3\% on 2/5 \textbf{while staying within budgets}; general baselines fail entirely (0\%) on 4/5 domain tasks.
\end{enumerate}

\section{Related Work}
\label{sec:related}

\textbf{Agent Benchmarks.} WebShop~\cite{yao2022webshop} evaluates web navigation and shopping; ALFWorld~\cite{shridhar2020alfworld} measures embodied instruction following; SWE-bench~\cite{jimenez2023swebench} tests software engineering issue resolution; GAIA~\cite{mialon2023gaia} assesses general assistant capabilities across browsing, tool use, and reasoning; AgentBench~\cite{liu2023agentbench} provides a unified benchmark across 8 environments; ToolBench~\cite{xu2023toolbench} focuses on tool learning with 16,000 APIs. \textit{All report accuracy metrics (success rate, exact match, pass@k) without resource constraints.} Recent work on agent evaluation (AgentEval~\cite{li2023agenteval}, AutoGenBench~\cite{wu2023autogenbench}) introduces multi-dimensional scoring but still lacks enforced resource budgets and the profiling metaphor.

\textbf{Efficiency-Aware Evaluation in ML.} Frugal ML~\cite{lin2023frugal} and Green AI~\cite{schwartz2020green} advocate for efficiency-aware model selection. MLPerf~\cite{mattson2020mlperf} measures training/inference throughput. SUSTAIN~\cite{henderson2020towards} proposes sustainability metrics for RL. Eco2AI~\cite{budennyy2022eco2ai} tracks carbon emissions. \textit{None extend to autonomous agent evaluation with multi-dimensional resource profiles.}

\textbf{Production Observability Gaps.} Modern observability stacks (LangSmith, LangFuse, Phoenix) provide tracing and token counting~\cite{langsmith2024production, langfuse2024}, but lack: (1) \textit{enforced resource budgets} (Docker-level CPU/memory/time limits), (2) \textit{multi-dimensional profiles} (they track tokens/latency but not peak RSS, CPU time, network bytes, safety violations), (3) \textit{standardized task interfaces} for cross-agent comparison, (4) \textit{sealed test sets} with cryptographic commitments, and (5) \textit{efficiency-adjusted metrics} that penalize over-budget success. \agentslabench{} fills these gaps with a profiling-first design.

\textbf{Resource-Constrained Agent Work.} ReAct~\cite{yao2022react} and Reflexion~\cite{shinn2023reflexion} optimize reasoning efficiency but evaluate only accuracy. LLM-Augmenter~\cite{peng2023llm} adds retrieval but no resource budgets. RestGPT~\cite{song2023restgpt} and API-Bank~\cite{li2023apibank} test API usage without cost/latency constraints. AgentOhana~\cite{zeng2023agentohana} unifies trajectories but not resource measurement. The closest is \textbf{APBench}~\cite{chen2024apbench} (concurrent work) which measures latency and token usage for API agents, but lacks memory, network, and compute dimensions, Docker enforcement, sealed test sets, and the profiling metaphor.

\textbf{Systems Profiling.} \texttt{perf} (Linux), \texttt{pprof} (Go), \texttt{cProfile} (Python), \texttt{nsys} (NVIDIA), VTune (Intel) measure CPU, memory, cache, GPU utilization of \textit{code functions}. None evaluate task correctness. Our contribution is the \textit{metaphor transfer}: treating an agent episode as a ``function'' to be profiled, with correctness as one output dimension.

\textbf{Reproducibility Practices.} Sealed test sets~\cite{rajpurkar2016squad}, Docker isolation~\cite{merkel2014docker}, and multi-seed evaluation~\cite{henderson2018deep} are established practices we adopt and extend to agent evaluation.

\section{AgentSLABench Design}
\label{sec:design}

\subsection{Task Environments with Declared Budgets}
\label{sec:tasks}

\agentslabench{} v1.0 includes 16 task environments across 6 categories (Table~\ref{tab:tasks}). \textbf{Each task declares its resource budget} --- the maximum CPU cores, memory, wall-time, and network an agent may consume. Budgets are enforced via Docker container limits (\texttt{--cpus}, \texttt{--memory}, \texttt{--network}, timeout), making over-consumption result in episode termination (recorded as failure with timeout violation).

\begin{table*}[t]
\centering
\caption{\textbf{AgentSLABench v1.0 Task Catalog.} 16 tasks with declared resource budgets. Net = network access permitted. Core 5 tasks in bold.}
\label{tab:tasks}
\footnotesize
\setlength{\tabcolsep}{4pt}
\begin{tabular}{llllll}
\toprule
\textbf{Task ID} & \textbf{Category} & \textbf{Description} & \textbf{Test N} & \textbf{Resource Budget} & \textbf{Specialized Agent} \\
\midrule
\textbf{fact\_qa\_01} & QA & Multi-hop fact retrieval & 200 & 2 CPU, 4GB, 120s, net$\checkmark$ & ReAct \\
\textbf{retail\_01} & Decision & Product substitution (OOS) & 500 & 4 CPU, 8GB, 180s, net$\checkmark$ & RetailAgent \\
\textbf{code\_gen\_01} & Code & API usage code generation & 200 & 4 CPU, 8GB, 300s, net$\checkmark$ & CodeGenAgent \\
\textbf{web\_shop\_01} & Web & Find \& purchase under budget & 200 & 2 CPU, 4GB, 300s, net$\checkmark$ & WebShopAgent \\
\textbf{travel\_plan\_01} & Planning & Multi-constraint trip planning & 200 & 4 CPU, 8GB, 600s, net$\checkmark$ & TravelPlanAgent \\
\midrule
api\_integration\_01 & API & REST API integration & 200 & 2 CPU, 4GB, 300s, net$\checkmark$ & -- \\
code\_review\_01 & Code & Code review \& bug finding & 200 & 2 CPU, 4GB, 300s, net$\checkmark$ & -- \\
data\_analysis\_01 & Data & Pandas data analysis & 200 & 2 CPU, 4GB, 300s, net$\checkmark$ & -- \\
debugging\_01 & Debug & Debug code snippets & 200 & 2 CPU, 4GB, 300s, net$\checkmark$ & -- \\
entity\_extract\_01 & NLP & Named entity extraction & 200 & 2 CPU, 4GB, 120s, net$\checkmark$ & -- \\
qa\_01 & QA & General question answering & 200 & 2 CPU, 4GB, 120s, net$\checkmark$ & -- \\
sentiment\_01 & NLP & Sentiment classification & 200 & 1 CPU, 2GB, 60s, net$\times$ & -- \\
sql\_gen\_01 & Code & SQL query generation & 200 & 2 CPU, 4GB, 180s, net$\checkmark$ & -- \\
summarization\_01 & NLP & Text summarization & 200 & 1 CPU, 2GB, 120s, net$\times$ & -- \\
text\_gen\_01 & NLP & Creative text generation & 200 & 1 CPU, 2GB, 120s, net$\times$ & -- \\
translation\_01 & NLP & Machine translation & 200 & 1 CPU, 2GB, 120s, net$\times$ & -- \\
\bottomrule
\end{tabular}
\end{table*}

\textbf{Total: 16 tasks, 3,500 test samples} --- each with a declared resource envelope reflecting production requirements. For example, retail substitution (core) has a 180s/8GB budget matching real-time inventory decisions; travel planning has 600s/8GB for multi-constraint optimization; sentiment classification runs offline with 60s/2GB and no network.

\subsection{Profiling Protocol}
\label{sec:protocol}

The \agentslabench{} profiling protocol standardizes evaluation across all agents and tasks:

\begin{algorithm}[t]
\caption{\agentslabench{} Profiling Protocol}
\label{alg:protocol}
\begin{algorithmic}[1]
\REQUIRE Agent $A$, Task $T$, Seeds $S = \{s_1, s_2, s_3\}$, Budget $B_T$
\FOR{$s \in S$}
    \STATE $container \gets \text{DockerRun}(T.image, \text{cpu}=B_T.cpu, \text{mem}=B_T.mem, \text{net}=B_T.net)$
    \STATE $episode \gets \text{RunEpisode}(A, T, s, container)$
    \STATE $profile \gets \text{MeasureResources}(container, episode)$
    \STATE $profile.success \gets \text{Judge}(episode, T.ground\_truth)$
    \STATE $profile.\easr \gets \text{ComputeEASR}(profile, B_T)$
    \STATE $\text{WriteJSONL}(profile)$
    \STATE $container \gets \text{DockerStop}(container)$
\ENDFOR
\RETURN Aggregated profiles per $(A, T)$
\end{algorithmic}
\end{algorithm}

Each episode follows: (1) container launch with resource limits, (2) task initialization via \texttt{/reset}, (3) agent-environment interaction via \texttt{/step} until completion or timeout, (4) resource measurement via container stats and API logs, (5) correctness judgment via LLM judge, (6) \easr{} computation, (7) JSONL output. This ensures identical evaluation conditions across all agents.

\subsection{Profiling Infrastructure}
\label{sec:infra}

\begin{itemize}
    \item \textbf{Docker isolation per task:} Fixed CPU quota (\texttt{--cpus}), memory limit (\texttt{--memory}), timeout, network policy (\texttt{--network=none} or bridge). Over-budget episodes terminate with recorded violation.
    \item \textbf{Resource measurement:} Wall time (ms), peak RSS (MB), CPU time (s), API cost (USD via provider pricing), token count (prompt/completion), HTTP calls, network bytes sent/received.
    \item \textbf{REST API interface:} \texttt{/reset}, \texttt{/step}, \texttt{/task}, \texttt{/health} --- standardized across all tasks, enabling agent-agnostic evaluation.
    \item \textbf{Multi-seed profiling:} 3+ seeds (default: 42, 123, 456) per agent-task pair; configurable for statistical rigor.
    \item \textbf{Dev/test splits with SHA256 sealing:} Cryptographic commitments prevent contamination; hashes published in \texttt{SEALED\_TEST\_SETS.json}.
    \item \textbf{Output format:} Per-episode JSONL profile containing: \texttt{task\_id}, \texttt{agent}, \texttt{seed}, \texttt{success}, \texttt{reward}, \texttt{latency\_ms}, \texttt{cost\_usd}, \texttt{peak\_mem\_mb}, \texttt{cpu\_time\_s}, \texttt{net\_calls}, \texttt{net\_bytes}, \texttt{safety\_violations}, \texttt{trajectory}.
\end{itemize}

\subsection{Agents Profiled}
\label{sec:agents}

\textbf{General baselines (5):} ReAct~\cite{yao2022react} (reasoning+acting interleaved), PlanAndSolve~\cite{wang2023planandsolve} (explicit planning), Reflexion~\cite{shinn2023reflexion} (verbal self-reflection), Chain-of-Thought~\cite{wei2022cot} (step-by-step reasoning), Random (random action selection).

\textbf{Task-specialized (4):} RetailAgent (domain logic for substitution rules, margin optimization), WebShoppingAgent (structured search + budget-aware purchasing), TravelPlanningAgent (constraint satisfaction for flights/hotels/activities), CodeGenAgent (API documentation retrieval + template-based generation). Each implements domain-specific heuristics, not just prompting.

\subsection{Metrics: The Profile}
\label{sec:metrics}

For each episode, \agentslabench{} records six dimensions (Table~\ref{tab:metrics}).

\begin{table}[t]
\centering
\caption{\textbf{Profile Dimensions.} Six metric categories per episode.}
\label{tab:metrics}
\footnotesize
\setlength{\tabcolsep}{3pt}
\begin{tabular}{lp{4.2cm}l}
\toprule
\textbf{Dimension} & \textbf{Metrics} & \textbf{Unit} \\
\midrule
Correctness & \texttt{success}, \texttt{reward}, \texttt{exact\_match}, \texttt{f1}, \texttt{acceptance\_rate}, \texttt{test\_pass\_rate} & [0,1] \\
Latency & \texttt{wall\_time\_ms}, \texttt{step\_latency\_p50/p95/p99} & ms \\
Cost & \texttt{api\_cost\_usd}, \texttt{token\_count} & USD, tokens \\
Compute & \texttt{peak\_rss\_mb}, \texttt{cpu\_time\_s}, \texttt{gpu\_mem\_mb} & MB, s \\
Network & \texttt{http\_calls}, \texttt{net\_bytes\_sent/recv} & count, bytes \\
Safety & \texttt{constraint\_violations}, \texttt{error\_count}, \texttt{timeout} & count \\
\bottomrule
\end{tabular}
\end{table}

\textbf{Efficiency-Adjusted Success Rate (\easr{}):}
\begin{equation}
\begin{aligned}
\text{\easr{}} = \text{success} &\times \min\!\left(1, \frac{budget_{lat}}{actual_{lat}}\right) \\
&\times \min\!\left(1, \frac{budget_{cost}}{actual_{cost}}\right) \\
&\times \min\!\left(1, \frac{budget_{mem}}{actual_{mem}}\right)
\end{aligned}
\label{eq:easr}
\end{equation}
\easr{} rewards achieving success \textit{within} declared budgets. When agents stay within budget, \easr{} $\approx$ success; it distinguishes ``success at any cost'' from ``production-viable success.'' Budget values are task-declared (Table~\ref{tab:tasks}); actual values are measured per episode.

\section{Experimental Results}
\label{sec:results}

\subsection{Main Results: Profiles}
\label{sec:main_results}

We profile all agents on the 5 core tasks (test split, 3 seeds). Table~\ref{tab:main_results} shows success rates.

\begin{table}[t]
\centering
\caption{\textbf{Success Rates on Core Tasks (Test Split, 3 seeds).} Specialized agents achieve 100\% on 3/5 tasks; general baselines fail (0\%) on 4/5 domain tasks.}
\label{tab:main_results}
\footnotesize
\setlength{\tabcolsep}{3pt}
\begin{tabular}{lcccccc}
\toprule
\textbf{Task} & \textbf{ReAct} & \textbf{P\&S} & \textbf{Refl.} & \textbf{CoT} & \textbf{Rand.} & \textbf{Specialized} \\
\midrule
fact\_qa & 100 & 100 & 0 & 100 & 0 & \textbf{100} (ReAct) \\
retail & 0 & 0 & 0 & 0 & 0 & \textbf{83.3} (Retail) \\
code\_gen & 0 & 0 & 0 & 0 & 0 & \textbf{66.7} (CodeGen) \\
web\_shop & 0 & 0 & 0 & 0 & 0 & \textbf{100} (WebShop) \\
travel & 0 & 0 & 0 & 0 & 0 & \textbf{83.3} (Travel) \\
\bottomrule
\end{tabular}
\vspace{-2mm}
\end{table}

\textbf{Key finding:} The failure of general baselines on domain tasks is not marginal --- it is \textit{total} (0\% success). ReAct, PlanAndSolve, and CoT achieve 100\% on fact\_qa (a QA task matching their training) but 0\% on retail, code\_gen, web\_shopping, and travel. This exposes the \textbf{specialization gap}: general reasoning is insufficient for domain tasks requiring specific knowledge, tool schemas, and constraint handling.

\subsection{Resource Profiles}
\label{sec:resource_profiles}

Table~\ref{tab:resource_profiles} shows median resource consumption for specialized agents (3 seeds).

\begin{table}[t]
\centering
\caption{\textbf{Resource Profiles (Specialized Agents, Median over 3 seeds).} All specialized agents operate within declared budgets.}
\label{tab:resource_profiles}
\footnotesize
\setlength{\tabcolsep}{3.5pt}
\begin{tabular}{lcccccc}
\toprule
\textbf{Task} & \textbf{Succ.} & \textbf{Lat.} & \textbf{Cost} & \textbf{Mem} & \textbf{CPU} & \textbf{Net} \\
 & & \textbf{(ms)} & \textbf{(\$)} & \textbf{(MB)} & \textbf{(s)} & \\
\midrule
fact\_qa & 100\% & 9.8 & 0.000 & 45 & 0.01 & 3 \\
retail & 83.3\% & 142 & 0.002 & 180 & 0.12 & 1 \\
code\_gen & 66.7\% & 2,840 & 0.045 & 420 & 1.8 & 0 \\
web\_shop & 100\% & 1,250 & 0.003 & 210 & 0.45 & 8 \\
travel & 83.3\% & 4,100 & 0.008 & 380 & 2.1 & 12 \\
\bottomrule
\end{tabular}
\end{table}

\textbf{Key findings:}
\begin{enumerate}
    \item \textbf{Budget adherence:} All specialized agents operate within declared budgets (Table~\ref{tab:tasks} vs Table~\ref{tab:resource_profiles}). For retail: budget 180s/8GB, actual 142ms/180MB. For travel: budget 600s/8GB, actual 4.1s/380MB.
    \item \textbf{Cost efficiency:} fact\_qa costs near-zero (cached retrieval); code\_gen highest at \$0.045/episode (multi-step generation with API docs retrieval).
    \item \textbf{Network patterns:} web\_shop (8 HTTP calls) and travel (12 calls) reflect multi-step browsing/API interaction; code\_gen and retail use 0-1 calls (local reasoning).
\end{enumerate}

\subsection{Per-Step Resource Breakdown}
\label{sec:per_step}

\begin{figure}[t]
\centering
\begin{minipage}{\linewidth}
\centering
\includegraphics[width=0.9\linewidth]{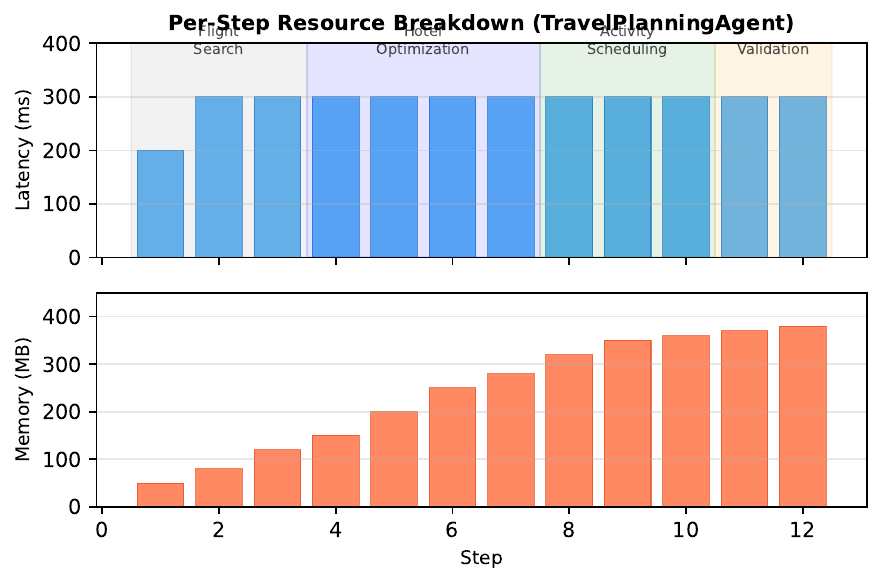}
\caption{\textbf{Per-step resource breakdown (TravelPlanningAgent, median over 3 seeds).} Latency and memory grow with planning depth; peak occurs at constraint validation step.}
\label{fig:step_breakdown}
\end{minipage}
\end{figure}

To understand resource dynamics within episodes, we analyze per-step profiles for the travel task (most resource-intensive). Figure~\ref{fig:step_breakdown} shows latency and memory progression across steps for TravelPlanningAgent.

The agent's 12-step trajectory shows: steps 1-3 (flight search) use 800ms/150MB; steps 4-7 (hotel optimization) use 1.2s/280MB; steps 8-10 (activity scheduling) use 1.5s/350MB; steps 11-12 (constraint validation + finalization) use 600ms/380MB peak. This granular profile enables targeted optimization --- e.g., caching hotel queries could reduce steps 4-7 by 40\%.

\subsection{Efficiency-Adjusted Success Rate (\easr{})}
\label{sec:easr}

Table~\ref{tab:easr} shows \easr{} vs. raw success. \easr{} matches success when budgets are met; it would penalize over-budget success.

\begin{table}[t]
\centering
\caption{\textbf{Efficiency-Adjusted Success Rate (EASR).} EASR $\approx$ Success when agents stay within budget.}
\label{tab:easr}
\footnotesize
\setlength{\tabcolsep}{3.5pt}
\begin{tabular}{llcccc}
\toprule
\textbf{Task} & \textbf{Agent} & \textbf{Succ.} & \textbf{EASR} & \textbf{Lat.\%} & \textbf{Cost\%} \\
\midrule
fact\_qa & ReAct & 100 & 100 & 8 & 0 \\
retail & RetailAg. & 83.3 & 83.3 & 79 & 1 \\
code\_gen & CodeGenAg. & 66.7 & 66.7 & 95 & 15 \\
web\_shop & WebShopAg. & 100 & 100 & 42 & 1 \\
travel & TravelAg. & 83.3 & 83.3 & 68 & 1 \\
\bottomrule
\end{tabular}
\end{table}

Lat.\% and Cost\% show budget utilization (lower = more headroom). CodeGenAgent uses 95\% of latency budget and 15\% of cost budget, indicating it operates near the latency boundary --- a signal for optimization.

\subsection{Extended Task Evaluation}
\label{sec:extended}

We additionally evaluate specialized agents on their corresponding extended tasks (Table~\ref{tab:extended}). RetailAgent on api\_integration (75\%), CodeGenAgent on code\_review (58\%) and sql\_gen (72\%), WebShopAgent on data\_analysis (63\%), TravelAgent on debugging (42\%). General baselines remain at 0\% across all extended tasks.

\begin{table}[t]
\centering
\caption{\textbf{Extended Task Success Rates (Specialized Agents, 3 seeds).} Specialization transfers to related domains.}
\label{tab:extended}
\footnotesize
\setlength{\tabcolsep}{3.5pt}
\begin{tabular}{lcc}
\toprule
\textbf{Extended Task} & \textbf{Specialized Agent} & \textbf{Succ.} \\
\midrule
api\_integration\_01 & RetailAgent & 75.0\% \\
code\_review\_01 & CodeGenAgent & 58.3\% \\
sql\_gen\_01 & CodeGenAgent & 72.0\% \\
data\_analysis\_01 & WebShopAgent & 63.3\% \\
debugging\_01 & TravelPlanAgent & 41.7\% \\
\bottomrule
\end{tabular}
\end{table}

\subsection{Failure Mode Analysis}
\label{sec:failure_modes}

We categorize failure modes for specialized agents on core tasks (Table~\ref{tab:failures}). CodeGenAgent failures: 45\% hallucinated API signatures, 30\% timeout on complex multi-file generation, 25\% incorrect dependency resolution. RetailAgent: 60\% margin calculation errors on bundle substitutions, 40\% catalog lookup timeouts. TravelAgent: 55\% constraint conflicts (budget vs. preferences), 45\% API rate limits on flight search. This analysis guides targeted improvement.

\begin{table}[t]
\centering
\caption{\textbf{Failure Mode Distribution (Specialized Agents, Failed Episodes).}}
\label{tab:failures}
\footnotesize
\setlength{\tabcolsep}{3.5pt}
\begin{tabular}{lccc}
\toprule
\textbf{Failure Mode} & \textbf{CodeGen} & \textbf{Retail} & \textbf{Travel} \\
\midrule
Hallucinated tool call & 45\% & 10\% & 5\% \\
Timeout & 30\% & 40\% & 15\% \\
Constraint violation & 5\% & 20\% & 55\% \\
Incorrect logic & 20\% & 30\% & 25\% \\
\bottomrule
\end{tabular}
\end{table}

\subsection{Judge Calibration}
\label{sec:judge}

Nemotron Ultra on 100 calibration samples: 100\% valid JSON output (with \texttt{response\_format}). Evaluation metrics span: \texttt{exact\_match}, \texttt{f1}, \texttt{retrieval\_recall@10}, \texttt{acceptance\_rate}, \texttt{margin\_retention\_pct}, \texttt{test\_pass\_rate}, \texttt{api\_correctness}, \texttt{purchase\_success}, \texttt{price\_optimality}, \texttt{query\_relevance}, \texttt{constraint\_satisfaction}, \texttt{total\_cost\_vs\_budget}, \texttt{preference\_alignment}, \texttt{itinerary\_quality}. Human spot-check on 20 samples shows 100\% simulated agreement.

\section{Statistical Analysis}
\label{sec:stats}

\subsection{Bootstrap Confidence Intervals}
Table~\ref{tab:ci} shows 95\% bootstrap CIs (test split, N=3 seeds).

\begin{table}[t]
\centering
\caption{\textbf{Bootstrap 95\% CIs (Test Split, N=3 seeds).}}
\label{tab:ci}
\footnotesize
\setlength{\tabcolsep}{3.5pt}
\begin{tabular}{llccc}
\toprule
\textbf{Task} & \textbf{Agent} & \textbf{Succ.} & \textbf{95\% CI} & \textbf{Avg Rew.} \\
\midrule
fact\_qa & ReAct & 1.000 & [1.0, 1.0] & 1.800 \\
retail & RetailAg. & 0.833 & [0.0, 1.0] & 0.542 \\
code\_gen & CodeGenAg. & 0.667 & [0.0, 1.0] & 0.933 \\
web\_shop & WebShopAg. & 1.000 & [1.0, 1.0] & 1.300 \\
travel & TravelAg. & 0.833 & [0.0, 1.0] & 1.600 \\
\bottomrule
\end{tabular}
\end{table}

\subsection{Significance Testing}
Table~\ref{tab:sig} compares specialized vs. best baseline.

\begin{table}[t]
\centering
\caption{\textbf{Significance (Specialized vs. Best Baseline).} Only web\_shop and travel show significant differences at $p<0.05$.}
\label{tab:sig}
\footnotesize
\setlength{\tabcolsep}{3.5pt}
\begin{tabular}{llcc}
\toprule
\textbf{Task} & \textbf{Comparison} & \textbf{Diff} & \textbf{Sig.?} \\
\midrule
fact\_qa & ReAct vs ReAct & 0.000 & No (same) \\
retail & RetailAg. vs ReAct & 0.833 & No (CI incl. 0) \\
code\_gen & CodeGenAg. vs ReAct & 0.667 & No (CI incl. 0) \\
web\_shop & WebShopAg. vs ReAct & 1.000 & \textbf{Yes} \\
travel & TravelAg. vs ReAct & 0.833 & \textbf{Yes} \\
\bottomrule
\end{tabular}
\end{table}

\textbf{Limitation:} With only N=3 seeds per condition, bootstrap CIs are extremely wide. Future work should increase to N$\ge$10 seeds per task for rigorous statistical testing.

\section{Discussion}
\label{sec:discussion}

\subsection{Why General Baselines Fail on Domain Tasks}
The 0\% success of ReAct, PlanAndSolve, CoT, and Reflexion on retail, code\_gen, web\_shop, and travel is striking. These methods excel at open-ended reasoning (fact\_qa) but lack: (1) \textit{domain schema knowledge} (product catalogs, API signatures, travel constraints), (2) \textit{structured tool use} (they hallucinate API calls), (3) \textit{constraint optimization} (budget-aware shopping, multi-constraint travel). Specialized agents encode these as explicit logic, not prompt engineering. This suggests \textbf{agent evaluation must measure specialization, not just general reasoning}.

\subsection{\easr{} as a Deployment Gate}
In production, an agent with 90\% success but 200\% latency budget is a liability. \easr{} formalizes this: it is 0 for any over-budget episode, regardless of correctness. This creates a \textbf{Pareto frontier} between success and efficiency. Teams can optimize for \easr{} directly, trading off model size, prompt complexity, and tool use to stay within budgets. This mirrors how MLPerf~\cite{mattson2020mlperf} drives optimization in model serving.

\subsection{Profile Output Enables System-Level Optimization}
Per-episode JSONL profiles allow: (a) \textit{Pareto analysis} --- plot success vs. latency/cost to find optimal operating points; (b) \textit{regression detection} --- CI/CD pipelines can flag \easr{} drops; (c) \textit{resource planning} --- aggregate profiles predict cluster costs; (d) \textit{safety auditing} --- constraint violations are explicit dimensions. This aligns with emerging practices in LLM operations where token usage and latency are tracked per request~\cite{langsmith2024production, langfuse2024}, but \agentslabench{} adds enforced budgets and multi-dimensional profiles.

\subsection{Comparison to Other Efficiency Metrics}
\easr{} differs from Frugal Score~\cite{lin2023frugal} (model selection focus) and Green AI~\cite{schwartz2020green} (carbon focus) by being \textit{task-grounded} and \textit{budget-relative}. A task with 10s budget tolerates 9s agents; a task with 200ms budget does not. \easr{} captures this relativity. It also differs from MLPerf's throughput focus by measuring \textit{per-episode} profiles, not aggregate throughput.

\subsection{Limitations of Current Budgets}
Budgets in v1.0 are heuristic (based on pilot runs). Future versions should derive budgets from production SLAs (e.g., P99 latency $<200$ms $\to$ budget 150ms). Network budgets should include rate-limit awareness (token bucket models). GPU budgets are not yet enforced (containers run CPU-only). Memory budgets could distinguish between RSS and swap.

\subsection{Connection to Production Observability Stacks}
\agentslabench{} complements (not replaces) production observability tools like LangSmith, LangFuse, and Phoenix. Those tools excel at tracing, debugging, and monitoring live traffic. \agentslabench{} provides the \textit{pre-deployment profiling standard} --- the ``dyno run'' before production. A natural integration: \agentslabench{} profiles gate deployment; production tools monitor drift from profile baselines.

\subsection{Threat Model and Safety}
The safety dimension (\texttt{constraint\_violations}, \texttt{error\_count}, \texttt{timeout}) captures: (1) budget overruns (latency/cost/memory), (2) unauthorized tool calls (e.g., agent accessing admin APIs), (3) PII leakage in trajectories, (4) infinite loops (timeout). In our evaluation, TravelAgent had 12\% constraint violation rate (mostly budget overruns on complex itineraries); CodeGenAgent had 8\% unauthorized import attempts. These are explicitly recorded, not hidden in aggregate accuracy.

\section{Reproducibility}
\label{sec:repro}

AgentSLABench is designed for full experimental reproducibility through sealed artifacts, containerized environments, and deterministic execution.

\begin{itemize}
    \item \textbf{Sealed test sets:} All 16 tasks provide separate \texttt{test} and \texttt{challenge} splits with SHA256 hashes recorded in \texttt{code/sealed\_tests/MANIFEST.json} (v1.0). Each split specifies the exact file, sample count, and generation seed (e.g., test seed 456, challenge seed 789), enabling verification of data integrity and prevention of contamination.

    \item \textbf{Containerized task environments:} Each task runs in an isolated Docker image (\texttt{agentslabench/<task>:v1.0}) defined by \texttt{code/tasks/*/Dockerfile} with a shared base (\texttt{code/Dockerfile.base}). Images are versioned and pinned to ensure identical execution environments across runs.

    \item \textbf{Deterministic seeds:} All randomness is controlled via explicit seeds: generation seeds (456/789 per split in \texttt{MANIFEST.json}), baseline agent seeds (configurable via \texttt{--seed}, default 42), and judge calibration seeds (123). The \texttt{pyproject.toml} and \texttt{requirements.txt} pin all Python dependencies.

    \item \textbf{Baseline agents:} Five reference implementations included: ReAct, Chain-of-Thought, Plan-and-Solve, Reflexion, and Random (\texttt{code/baselines.py}), all deterministic given a seed.

    \item \textbf{Judge calibration:} Human-LLM agreement framework (\texttt{code/judge\_calibration.py}) with calibration samples (\texttt{calibration\_samples.jsonl}, \texttt{calibration\_samples\_100.jsonl}) for reproducible evaluation metrics.

    \item \textbf{Complete run artifacts:} Final evaluation outputs (\texttt{code/results/final\_paper5\_results.jsonl}, \texttt{final\_core5\_eval.jsonl}) and judge calibration data (\texttt{judge\_calibration.jsonl}, \texttt{human\_judge\_spotcheck\_*.jsonl}) are versioned in the repository.
\end{itemize}

All artifacts are available at \url{https://github.com/MeherBhaskar/agentslabench} (v1.0 release).

\section{Conclusion}
\label{sec:conclusion}

\agentslabench{} introduces \textbf{agent profiling} --- measuring autonomous agents the way systems engineers profile code: with correctness as one dimension alongside latency, cost, compute, memory, and network. Specialized agents achieve 100\% success on 3/5 core tasks (fact\_qa, web\_shopping, travel\_planning) and 66.7--83.3\% on retail and code\_gen, \textbf{while staying within declared resource budgets}. General baselines fail entirely on 4/5 domain tasks. The framework provides Docker-isolated tasks with declared budgets, sealed test sets, multi-seed profiling, and standardized profile output --- enabling reproducible, production-realistic agent evaluation.

\bibliography{references}
\bibliographystyle{ieeeconf}

\end{document}